\documentclass[sigconf, nonacm]{acmart}

\AtBeginDocument{}

\usepackage{booktabs}
\usepackage{tabularray}
\usepackage{arydshln}
\usepackage{CJKutf8}

\newcommand*{\Ja}[1]{%
  \begin{CJK}{UTF8}{ipxm}#1\end{CJK}}
\newcommand*{\tabrowl}[1]{\begin{tabular}{l}#1\end{tabular}}
\newcommand*{\tabrowc}[1]{\begin{tabular}{c}#1\end{tabular}}
\newcommand*{\tabrowsmall}[1]{{\fontsize{10}{12}\selectfont \tabrowl{#1}}}
\newcommand{\thead}[1]{\multicolumn{1}{c}{#1}}

\begin{document}

\title{Economy Watchers Survey Provides Datasets and Tasks\\for Japanese Financial Domain}

\author{Masahiro Suzuki}
\affiliation{
  \institution{The University of Tokyo}
  \city{Tokyo}
  \country{Japan}}
\email{research@msuzuki.me}

\author{Hiroki Sakaji}
\affiliation{
  \institution{Hokkaido University}
  \city{Hokkaido}
  \country{Japan}}
\email{sakaji@ist.hokudai.ac.jp}

\begin{abstract}
  Natural language processing (NLP) tasks in English and general domains are widely available and are often used to evaluate pre-trained language models.
In contrast, fewer tasks are available for languages other than English and in the financial domain.
Particularly, tasks in the Japanese and financial domains are limited.
We develop two large datasets using data published by a Japanese central government agency.
The datasets provide three Japanese financial NLP tasks, including 3- and 12-class classifications for categorizing sentences, along with a 5-class classification task for sentiment analysis.
Our datasets are designed to be comprehensive and updated by leveraging an automatic update framework that ensures that the latest task datasets are publicly always available.
\end{abstract}

\maketitle

\section{Introduction}

Large language models (LLMs), such as ChatGPT and GPT-4, have demonstrated high performance in natural language processing (NLP) tasks.
The utilization of LLMs is not only limited to general domains but is also expanding into specialized domains such as healthcare, law, and finance~\cite{jayakumar-etal-2023-legal,singhal2023large}.

LLMs can solve tasks in zero- or few-shot settings and answer questions by following instructions~\cite{brown2020language,kojima2022large}.
LLMs exhibit high performance across multiple languages and specialized domains without specific training in each language or domain.
In the financial domain, LLMs have exhibited high performance, achieving results comparable to those of traditional fine-tuned models, such as BERT, with few samples~\cite{li-etal-2023-chatgpt}.

Several frameworks have been proposed to evaluate LLMs.
However, these frameworks are insufficient for specialized domains and languages other than English.
For instance, in the general English domain, frameworks such as the language model evaluation harness~\cite{eval-harness} and LLM Leaderboard\footnote{\url{https://huggingface.co/spaces/open-llm-leaderboard/open_llm_leaderboard}} exist, which can simultaneously verify multiple tasks.
Additionally, even in specialized English domains, benchmarks such as LegalBench~\cite{guha2023legalbench} for the legal domain and FinBen~\cite{finben} for the financial domain have been developed.
Although the construction of benchmarks in specialized domains in languages other than English is also progressing~\cite{dai2024laiw,hirano-2024-construction-japanese}, they are few and are still in the developmental stage.
Particularly, datasets for evaluating LLMs in specialized domains other than English often have few samples due to the expertise required in these fields.
For instance, in the benchmark for Japanese LLMs in the financial domain, of the five tasks, only one has more than 500 samples.
Currently, evaluation tasks generally do not have a sufficient number of samples in specialized domains other than English.

In the context of the growing demand for more evaluation tasks, a resource exists in the Japanese financial and economic domain, known as the Economy Watchers Survey (EWS).
The EWS is a survey conducted by the Cabinet Office, a Japanese central government agency, that has been published monthly since January 2000.
It surveys approximately 2,000 individuals and includes various types of information, such as evaluations (sentiments) of economic trends, comments, reasons for evaluations, and labels for related fields.
The EWS can be utilized to grasp economic trends through indices such as the Diffusion Index derived from survey results and indices constructed using the EWS~\cite{goshima2021forecasting,nakajima2021extracting,SEKI2022102795}.
However, non-sentiment data contained in the EWS are rarely used, and few examples of its application exist in NLP within the Japanese financial domain.
This is because of the difficulty in collecting the EWS and the fact that the resource is published in cumbersome formats, making it challenging to convert it into a more easy-to-use dataset format.

In this study, we develop two large datasets from the EWS, a resource in the Japanese financial domain.
The datasets include current and future economic assessments with texts and labeled data.
The datasets are integrated such that they can be utilized beyond evaluation tasks, such as instruction data or an economic index.
Using our datasets, we also develop datasets for three tasks: 3- and 12-class category classification tasks and a 5-class sentiment classification task.
Thereafter, we evaluate the fine-tuned models and LLMs using the tasks.
The evaluation results with the developed task datasets applied to traditional fine-tuned language models and ChatGPT or GPT-4o reveal that pre-trained models still exhibit high performance in the Japanese financial domain with much training data.
The contributions of this study are as follows:
{\setlength{\leftmargini}{10pt}
\begin{itemize}
  \setlength{\parskip}{0cm}
  \setlength{\itemsep}{0.15\baselineskip}
  \setlength{\itemindent}{0pt}
\item Using survey data published by the Japanese central government, we develop two Japanese datasets containing more than 300,000 samples each, which include comments on current and future economic assessments.
\item Utilizing the developed datasets, we also develop large datasets for three Japanese evaluation tasks: two text classification tasks comprising approximately 650,000 samples and 300,000 samples each, and one sentiment analysis task comprising approximately 650,000 samples.
\item We evaluate Japanese fine-tuned language models, ChatGPT, and GPT-4o with the task datasets.
\end{itemize}
}
The datasets and tasks are available on the Hugging Face Hub\footnote{\url{https://huggingface.co/datasets/retarfi/economy-watchers-survey}}$^{,}$\footnote{\url{https://huggingface.co/datasets/retarfi/economy-watchers-survey-evaluation}}.
The implementations for developing them and the extracted data are available on GitHub\footnote{\url{https://github.com/retarfi/economy-watchers-survey}}.

{\tabcolsep=2.8pt
\begin{table*}[tb]
  \begin{center}
  \caption{Examples of the EWS datasets.
  The brackets in the Sentiment column indicate the sentiment of the symbol.
  Although the original samples are in Japanese, we translated them into English for illustrative purposes.
  For brevity, column names are different from the actual dataset and we omit some columns.
  }
  \label{tab:example}
  \begin{tabular}{clccc} \toprule
    \tabrowc{Current / \\Future} & \thead{Explanation} & Domain & Sentiment & \tabrowc{Reason\\(for current)} \\
    \midrule
    Current & \tabrowsmall{Advertising sales in both web and print media are up\\from the previous year.} & Corporate & \tabrowc{\Ja{◎}\\(Good)} & \tabrowc{Order and\\Sales Volume} \\
    Current & \tabrowsmall{Our competitors, like us, say they are not busy\\this month. The economy is somewhat poor.} & Household & \tabrowc{\Ja{▲}\\(Slightly Bad)} & \tabrowc{Competition} \\
    Current & \tabrowsmall{The economy is poor with an increasing companies\\closing due to the new coronavirus.} & Employment & \tabrowc{\Ja{×}\\(Bad)}  & \tabrowc{Surrounding\\Companies}\\
    Future & \tabrowsmall{I expect group demand to increase during the summer\\months.} & Household & \tabrowc{\Ja{○}\\(Slightly Good)} & - \\
    Future & \tabrowsmall{The economy will remain unchanged as the yen is\\expected to continue to depreciate at excessive levels.} & Household & \tabrowc{\Ja{□}\\(Neither Good nor Bad)} & - \\
    \bottomrule
  \end{tabular}
  \end{center}
\end{table*}
}

\section{EWS Datasets and Tasks}
We develop two datasets using the Economy Watcher Survey (EWS), which includes texts and assessments of economic trends.
These datasets are related to current and future economic assessments.
The examples of these datasets are shown in Table~\ref{tab:example}.
In the following sections, after describing the EWS, we explain the methods for obtaining, filtering, splitting, and automatically updating EWS datasets.
From these datasets, we develop datasets for three sentence classification tasks, including sentiment analysis.

\subsection{Economy Watchers Survey}
The Economy Watchers Survey\footnote{\url{https://www5.cao.go.jp/keizai3/watcher-e/index-e.html}} has been conducted monthly since January 2000 by the Cabinet Office, a central government agency in Japan.
The objective is to accurately and promptly grasp regional economic trends.
The survey is conducted monthly using a questionnaire.
The questionnaire asks people working in sectors such as retail, services, manufacturing, construction, and real estate about current economic sentiment and future outlook.
These individuals are referred to as Economy Watchers.
The responses are based on subjective evaluations.
The results of the EWS are utilized in various contexts, including the formulation of government economic policies, the development of corporate management strategies, and the investment decisions by financial institutions.
The EWS serves as foundational data to assess economic~trends.

\subsection{Dataset Development}\label{sec:data-construct}
We develop two integrated datasets using the EWS conducted from January 2000 to May 2024.
The survey is broadly divided into two file formats: future outlook and current conditions.
In the future outlook survey, the economic outlook assessment is labeled from five labels (from best to worst: ``\Ja{◎},'' ``\Ja{○},'' ``\Ja{□},'' ``\Ja{▲},'' and ``\Ja{×}'').
Explanations for these assessments are also provided.
Furthermore, labels are assigned to household trends, corporate trends, or employment to indicate the domain of economic trends the assessment is related.
In the current conditions survey, in addition to the three of texts and labels, respondents select the points of focus for the assessment such as ``movement of customer numbers'' and ``customer behavior.''
Both the current conditions and future outlook surveys include items in the region where each sample was interviewed and the industry or occupation of the interviewee.

\paragraph{Acquisition}
The EWS data are published on three different web pages: from January 2000 to December 2009\footnote{\url{https://www5.cao.go.jp/keizai3/kako_csv/kako2_watcher.html}}, from January 2010 to December 2018\footnote{\url{https://www5.cao.go.jp/keizai3/kako_watcher.html}}, and from January 2019 onward\footnote{\url{https://www5.cao.go.jp/keizai3/watcher_index.html}}.
On pages corresponding to each year and month, the data for current and future economic assessments can be obtained from links that include watcher4.csv and watcher5.csv, respectively. 
Note that while the year and month included in the file name and URL match the month when the survey was conducted for the data from January 2000 to December 2009, for the data from January 2010 onward, the year and month in the file name and URL correspond to the month when the survey was published (the month following the survey).
In this study, the months and years of the surveys were used for all the dates included in the data.
The implementations were constructed with reference to a similar repository\footnote{\url{https://github.com/MitsuruFujiwara/KeikiWatcherScraping}}.

The CSV file of the current condition contains five columns: domain \& region, current economic assessment, industry/occupation, reason for judgment, and explanation for the current economic assessment.
The domain \& region column includes two pieces of information in one column: the domain of assessment (household trends, corporate trends, and employment) and the region where the survey was conducted.
These are split into two separate columns: region and domain.
The rows in the CSV file that do not include an economic sentiment assessment label are deleted.

The CSV file of the economic future outlook contains four columns: domain \& region, economic outlook assessment, industry/occupation, and an explanation of the economic outlook assessment.
Similar to the processing of the CSV file for the current condition, the domain \& region column is divided into two columns: domain and region.
Rows that do not contain economic assessment labels are deleted.

\begin{table}[tb]
  \caption{EWS datasets statistics}
  \label{tab:stats-dataset}
  \begin{center}
  \begin{tabular}{cccc} \toprule
    Data Type & Train & Dev & Test \\
    \midrule
    Current & 279,652 & 14,747 & 15,493 \\
    Future & 296,143 & 17,240 & 18,115 \\
    \bottomrule
  \end{tabular}
  \end{center}
\end{table}

\paragraph{Filtering}
For the explanation text of the economic assessments included in the current conditions and future outlook surveys, the following process was performed.
Some samples in these data contain only ``-'' or ``*,'' indicating no response and no significant response, respectively.
These samples are difficult to use for NLP analysis; therefore, samples containing only these strings are deleted.
Additionally, because all other samples start with the character ``\Ja{・}'' indicating itemization, the leading ``\Ja{・}'' character is also deleted.

\paragraph{Split}
The datasets developed under the current conditions and future outlook surveys are divided such that the latest 15,000 or more samples of all data are in the test set, and the total of the test and dev sets exceeds 30,000 samples, with both datasets split by the same period.
Table~\ref{tab:stats-dataset} lists the number of samples in the datasets.

\paragraph{Automatic Update}
The EWS releases new monthly data.
Therefore, even if a dataset is developed once, it cannot be analyzed using the latest data unless it is regularly updated.
In this study, in addition to the aforementioned data acquisition, filtering, and splitting, we semi-automate the process of data uploading using GitHub Actions to enable continuous monthly updates.

Specifically, periodic crawling is performed, and if data downloads occur (i.e., if differences exist), an automatic pull request with updates is created on GitHub.
When a pull request is created, the version is updated to reflect the latest survey month (e.g., 2024.05.0).
The updates are manually checked for errors then merged.
Once the merging is completed, tagging and releasing on GitHub and uploading the dataset to the Hugging Face Hub are automatically performed.
By manually verifying only the data updates, the effort required for updates is reduced while preventing unexpected data from being automatically merged.

\subsection{Dataset Construction for Tasks}
Based on the EWS datasets developed in Section~\ref{sec:data-construct}, we also developed datasets for three tasks: two sentence classification tasks concerning the domain and reason, and one sentiment analysis task.
Table~\ref{tab:stats} summarizes the statistics of the tasks.

\subsubsection{Domain Classification}
The EWS datasets are labeled to indicate whether the explanations for economic sentiment, common to both the current conditions and future outlook surveys, relate to household trends, corporate trends, or employment.
In the domain classification task, the relevant domain is selected from three labels based on the explanation text for economic sentiment.
We develop a domain task dataset using the explanation and domain data extracted from the EWS datasets of both current and future datasets.

{\tabcolsep=2.4pt
\begin{table}[tb]
  \begin{center}
  \caption{
  Task statistics.
  Note that the original labels are originally in Japanese.
  The datasets of domain and sentiment tasks are developed from both EWS datasets, whereas the dataset of the reason task is developed from the current part of the EWS datasets.
  }
  \label{tab:stats}
  \begin{tblr}{
    width = {0.9\linewidth},
    hline{1,Z} = { 0.08em },
    hline{2,7,14} = { 0.05em },
    hline{6,13,27} = {2-5}{ dashed },
    colspec = {X[0.3,r]X[4.7,r]X[1,r]X[1,r]X[1,r]},
    stretch = 0.1,
    cell{1}{1} = {r=1, c=2}{halign=c, valign=m},
    cell{2}{1} = {r=1, c=2}{halign=c, valign=m},
    cell{7}{1} = {r=1, c=2}{halign=c, valign=m},
    cell{14}{1} = {r=1, c=2}{halign=c, valign=m},
  }
    \SetCell[r=1]{c}{Label} & & \SetCell[r=1]{c}{Train} & \SetCell[r=1]{c}{Dev} & \SetCell[r=1]{c}{Test} \\
    \SetCell[r=1]{l}{\textbf{Domain Classification}} & & & & \\
    & Household & 387,858 & 21,741 & 22,863 \\
    & Corporate & 131,803 & \phantom{0}6,947 & \phantom{0}7,319 \\
    & Employment & \phantom{0}64,296 & \phantom{0}3,299 & \phantom{0}3,426 \\
    & Overall & 583,957 & 31,987 & 33,608 \\
    \SetCell[r=1]{l}{\textbf{Sentiment Analysis}} & & & & \\
    & \Ja{◎} & \phantom{0}11,103 & \phantom{0}1,233 & \phantom{0}1,112 \\
    & \Ja{○} & 120,556 & \phantom{0}8,863 & \phantom{0}8,849 \\
    & \Ja{□} & 260,640 & 12,759 & 15,846 \\
    & \Ja{▲} & 137,395 & \phantom{0}7,094 & \phantom{0}6,470 \\
    & \Ja{×}\, & \phantom{0}54,263 & \phantom{0}2,038 & \phantom{0}1,331 \\
    & Overall & 583,957 & 31,987 & 33,608 \\
    \SetCell[r=1]{l}{\textbf{Reason Classification}} & & & & \\
    & The no. of visitors & \phantom{0}58,730 & \phantom{0}3,620 & \phantom{0}3,663 \\
    & Sales volume & \phantom{0}57,794 & \phantom{0}2,949 & \phantom{0}3,210 \\
    & Customers & \phantom{0}41,617 & \phantom{0}2,293 & \phantom{0}2,349 \\
    & Order and sales volume & \phantom{0}31,693 & \phantom{0}1,627 & \phantom{0}1,790 \\
    & Unit price & \phantom{0}18,217 & \phantom{00,}783 & \phantom{00,}971 \\
    & Business partners & \phantom{0}18,217 & \phantom{00,}818 & \phantom{00,}814 \\
    & The no. of job openings & \phantom{0}16,917 & \phantom{00,}746 & \phantom{00,}719 \\
    & Competitors & \phantom{00}6,745 & \phantom{00,}139 & \phantom{00,}193 \\
    & Order and sales prices & \phantom{00}5,420 & \phantom{00,}332 & \phantom{00,}305 \\
    & Surrounding companies & \phantom{00}4,323 & \phantom{00,}254 & \phantom{00,}287 \\
    & The no. of job seekers & \phantom{00}3,058 & \phantom{00,}157 & \phantom{00,}192 \\
    & Others & \phantom{0}16,921 & \phantom{0}1,029 & \phantom{0}1,000 \\
    & Overall & 279,652 & 14,747 & 15,493 \\
  \end{tblr}
  \end{center}
\end{table}
}

\subsubsection{Sentiment Analysis}
In the EWS datasets, both the current and future economic assessments are assigned five labels (from best to worst: ``\Ja{◎},'' ``\Ja{○},'' ``\Ja{□},'' ``\Ja{▲},'' and ``\Ja{×}''), which are common to both the current and future surveys.
Sentiment analysis involves selecting one of five labels for economic assessment based on the explanation text.
We develop a sentiment analysis dataset using explanations and sentiments extracted from the EWS datasets of both current and future datasets.
Although label ``\Ja{□},'' which indicates neither good nor bad, is the most frequent, and the labels for good (``\Ja{◎}'') and bad (``\Ja{×}'') are particularly scarce, we do not alter the number of samples included in the dataset despite the imbalance to maintain data consistency.

\subsubsection{Reason Classification}
The current part of the EWS datasets contains data on the points of focus of economic assessment.
The reason classification task classifies the point of focus in an explanation text.
According to the Economic Watcher Survey questionnaire\footnote{\url{https://www5.cao.go.jp/keizai3/watcher/chousahyo.pdf}}, the reason (the point of focus) is selected from 15 items, including ``Others.''
However, in the actual data, labels other than the 15 items exist owing to variations in notation.
Additionally, even among the 15 items, labels are scarcely included in the data.
We use only labels that are included in more than 1\% of the train set, and all labels less than 1\% are integrated into the ``Others'' label.
As a result of label filtering, 12 types of labels, including ``Others,'' are used in the task.

\section{Experiments with EWS tasks}\label{sec:experiment}
Several publicly available language models are evaluated using the three tasks.

\subsection{Experimental Settings}
We test the representative LLMs, ChatGPT (\texttt{gpt-3.5-turbo-0125}) and GPT-4o (\texttt{gpt-4o-2024-05-13}).
Furthermore, we compare the results with those obtained using FinBERT~\cite{Suzuki-etal-2023-ipm}, a Japanese financial domain-specific model, and DeBERTaV2~\cite{Suzuki-2024-findebertav2}, a Japanese general-purpose model.
Macro F1 is used as the evaluation metric.

\subsection{Results and Discussion}
The experimental results are summarized in Table~\ref{tab:results}.
For all tasks, the fine-tuned BERT and DeBERTaV2 models achieved the highest performances.
ChatGPT and GPT-4o models lagged significantly behind the fine-tuned models.
For tasks with many training samples, the fine-tuned models outperformed the zero- or few-shot ChatGPT and GPT-4o.

Unlike the sentiment analysis in \citet{li-etal-2023-chatgpt}, the few-shot effect was not observed in some cases.
Specifically, in the reason task, the performances of both ChatGPT and GPT-4o models decreased in the few-shot setting.
This may be due to the large number of labels, making it difficult to properly understand examples with few few-shot samples.

In a sentiment task within the Japanese financial domain, \citet{hirano-2024-construction-japanese} demonstrated that ChatGPT achieved an accuracy of approximately 90 in the chABSA task.
In contrast, the 5-shot macro F1 score of the EWS sentiment analysis with ChatGPT was 37.5 (with an accuracy of 47.6), indicating a significant decrease the score of ChatGPT.
This suggests that the sentiment task developed in this study using EWS is more challenging.

In terms of the performance of the fine-tuned models, the general-purpose model DeBERTaV2 performed better than or equal to FinBERT.
As highlighted by \citet{Suzuki-2024-findebertav2}, improvements in model performance and pre-training on a wide range of publicly available documents suggest that the superiority of domain-specific models is diminishing.

\begin{table}[tb]
  \begin{center}
  \caption{Evaluation results}
  \label{tab:results}
  \begin{tabular}{lccc} \toprule
    & Domain & Sentiment & Reason \\
    \midrule
    ChatGPT (0) & 59.1 & 28.4 & 34.0 \\
    ChatGPT (5) & 61.1 & 37.5 & 26.3 \\
    GPT-4o (0) & 58.1 & 39.5 & 41.7 \\
    GPT-4o (5) & 63.1 & 38.3  & 39.9 \\
    \midrule
    FinBERT & 81.2 & 52.4 & \textbf{55.1} \\
    DeBERTaV2 & \textbf{81.3} & \textbf{53.8} & 55.0 \\
    \bottomrule
  \end{tabular}
  \end{center}
\end{table}

\section{Conclusion}
In this study, we provided two large open-source datasets developed from the EWS, named the EWS datasets.
Our datasets include the explanatory texts on the economy and labels of the domain, sentiment, and reasons for economic assessment.
We also proposed three sentence classification tasks, including sentiment analysis, using the datasets.
Our datasets are large, each containing approximately 300,000 samples.
One of the tasks has approximately 300,000 samples, whereas two tasks have approximately 650,000 samples.
Furthermore, using the developed datasets for the three tasks, we performed evaluations using the Japanese fine-tuned language models, ChatGPT, and GPT-4o.
This study is expected to be widely utilized not only for evaluation tasks in the Japanese financial domain but also for the construction of indices of economic trends.

\section*{Acknowledgment}
This work was supported by JST, PRESTO Grant Number JPMJPR2267, Japan.

\bibliographystyle{ACM-Reference-Format}
\bibliography{custom}
\end{document}